# Improving Uyghur ASR systems with decoders using morpheme-based language models

Zicheng Qiu[a,1], Wei Jiang[a], Turghunjan Mamut[a]

[a]*College of Information Engineering, Tarim University, Alar City, Xinjiang, 843300, P.R. China*

**Abstract:** Uyghur is a minority language, and its resources for Automatic Speech Recognition (ASR) research are always insufficient. THUYG-20 is currently the only open-sourced dataset of Uyghur speeches. State-of-the-art results of its clean and noiseless speech test task haven't been updated since the first release, which shows a big gap in the development of ASR between mainstream languages and Uyghur. In this paper, we try to bridge the gap by ultimately optimizing the ASR systems, and by developing a morpheme-based decoder, MLDG-Decoder (Morpheme Lattice Dynamically Generating Decoder for Uyghur DNN-HMM systems), which has long been missing. We have open-sourced the decoder. The MLDG-Decoder employs an algorithm, named as "on-the-fly composition with FEBABOS", to allow the back-off states and transitions to play the role of a relay station in on-the-fly composition. The algorithm empowers the dynamically generated graph to constrain the morpheme sequences in the lattices as effectively as the static and fully composed graph does when a 4-Gram morpheme-based Language Model (LM) is used. We have trained deeper and wider neural network acoustic models, and experimented with three kinds of decoding schemes. The experimental results show that the decoding based on the static and fully composed graph reduces state-of-the-art Word Error Rate (WER) on the clean and noiseless speech test task in THUYG-20 to 14.24%. The MLDG-Decoder reduces the WER to 14.54% while keeping the memory consumption reasonable. Based on the open-sourced MLDG-Decoder, readers can easily reproduce the experimental results in this paper.

## 1. Introduction

In recent years, Large Vocabulary Continuous Speech Recognition (LVCSR) technology, applied to mainstream languages such as English[1][2] and Chinese[3], has made significant progress. The development has been triggered and accelerated by the Deep Learning (DL) technologies [1][2] that can learn rules and knowledge about pronunciation, vocabulary, syntax and pragmatic from a large amount of training data[4][5]. However, Uyghur is a minority language, and resources available for training a Uyghur Automatic Speech Recognition (ASR) system are relatively scarce [6-13], which hinders the newly emerging DL technologies from being applied to Uyghur ASR.

Due to DNN-HMM (Deep Neural Network-Hidden Markov Model) hybrid system's revival, the performance of English ASR early-bird systems leapfrogs [1][2]. In the early days, feed-forward DNNs were firstly tried. Subsequently, Recurrent Neural Networks (RNNs) [14], Convolutional Neural Networks (CNNs) [15], Long Short-Term Memory units (LSTMs) [16] and Time Delay Neural Networks (TDNNs) [17] were used to exploit more contextual information to build more powerful English acoustic models. The English DNN-HMM ASR systems have achieved human parity on conversational speech recognition tasks [18]. Recently, it has become the mainstream to construct sequence-to-sequence models by end-to-end training, while attention mechanism is

---

[1] Corresponding author

Email address: **zichengqiu@gmail.com** (Zicheng Qiu)

mainly concerned [19]. The end-to-end approaches, for the ASR systems of mainstream languages, have shown competitive performance to the state-of-the-art hybrid approaches [4].

Meanwhile, the newly emerging ASR technologies applied to the mainstream languages are being introduced to Uyghur. However, the progress of Uyghur ASR is not significant. Some of Uyghur ASR research uses private speech datasets. Feed-forward DNNs [7-9] and LSTMs [10] are applied to train Uyghur acoustic models, but the experiments are all based on their own private speech datasets, which makes the results not comparable. Other Uyghur ASR research uses THUYG-20 [6], which is the only open-sourced Uyghur speech dataset, along with which a package of Kaldi [20] recipes is released[2]. THUYG-20 has been included in the well-known resource website OpenSLR[3]. When THUYG-20 was firstly released, the acoustic model was constructed by a DNN with 4 hidden layers, and two morpheme-based Language Models (LM) helped it to achieve state-of-the-art WERs [6]. After that, based on THUYG-20, a DNN acoustic model with 6 hidden layers has been trained, while a CNN with 2 convolutional layers and 4 fully connected layers has also been trained and outperformed the DNN acoustic model [11]. Some more optimization [12] and transfer learning [13] have been applied to enhance the performance on THUYG-20. However, all the efforts have failed to outperform the WER firstly released and published along with THUYG-20 in [6]. One important reason is that the morpheme-based decoding for the DNN-HMM systems has been disabled due to the lack of a decoder based on on-the-fly composition algorithm. The missing morpheme-based decoder makes it impossible for researchers to reproduce the experimental results in [6], and also becomes an obstacle for the newly emerging technologies to be applied to THUYG-20, which helps to build a big gap between mainstream languages and Uyghur in the development of ASR. Additionally, most of the relevant studies on Uyghur ASR are all published in Chinese and their influence is limited [6-8][10-13].

To bridge the gap, transfer learning is widely used to build Uyghur acoustic models, by borrowing training data from mainstream languages [13][21][22]. A project, which is named as "Multilingual Minorlingual Automatic Speech Recognition (M2ASR)" [21], has received much attention. The project is dedicated to build ASR systems for 5 minority languages in China, where Uyghur is included. In the published research work, DNNs are trained with data of a rich-resource language by employing the transfer learning approach, and then acoustic models for minority languages are constructed by relabeling the targets in a semi-supervised learning phase [22]. In [13], the Uyghur acoustic model with 6 hidden layers is trained by using multi-language shared hidden layers, based on THUYG-20.

It is another hot topic to optimize the set of sub-word units for the lexicon and the LM according to the agglutinative property of Uyghur [9][23][24]. Modeling based on word units will lead to an explosion of the vocabulary, and low lexicon coverage. Morpheme units are usually adopted to keep sufficient coverage, but many short morphemes will be in confusion due to co-articulation [9][24]. An optimized set of modeling units is built by selecting the best one between the morpheme and word units to retain both the merits of the two [9][23]. In [24], words are split into stems and suffixes to build lexicons and LMs with sub-word units.

Despite of all the efforts, state-of-the-art results of THUYG-20 haven't been updated since the first release and publication. In this paper, we ultimately optimize the training process of the GMM-HMM and DNN-HMM systems successively, which makes it possible to train deeper and wider neural network acoustic models. On the other hand, we independently develop a morpheme-based decoder for the DNN-HMM system, based on the Kaldi platform[20]. We have disclosed this decoder's source code, so that readers can easily reproduce the experimental results in this paper based on THUYG-20.

---

[2] https://github.com/wangdong99/kaldi/blob/master/egs/thuyg20/s5

[3] https://openslr.org/22/

The remainder of this paper is organized as follows. In section 2, THUYG-20 speech dataset is briefly introduced. In Section 3, the GMM-HMM baseline systems are introduced and optimization schemes of them are figured out. In section 4, deeper and wider DNNs are introduced to construct acoustic models, and the development of a morpheme-based decoder for the DNN-HMM systems is introduced. In this section, the algorithm of on-the-fly composition is analyzed in detail. In section 5, the experimental results of the GMM-HMM and DNN-HMM systems are shown respectively. Finally, we conclude in section 6.

## 2. An open/free corpus of read Uyghur speech: THUYG-20

**Table 1 The phoneme set annotated by Latin letters in THUYG-20**

| Phoneme | Latin | Phoneme | Latin | Phoneme | Latin |
|---|---|---|---|---|---|
| ئ | v | ف | f | ت | t |
| **Sil** | sil | گ | g | ئۇ | u |
| ئە | A | ھ | h | ئە(emza) | vA |
| غ | G | ئى | i | ئۆ(emza) | vO |
| خ | H | ج | j | ئۈ(emza) | vU |
| ڭ | J | ك | k | ئا(emza) | va |
| ن | N | ل | l | ئې(emza) | ve |
| ئۆ | O | م | m | ئى(emza) | vi |
| ئۈ | U | ن | n | ئو(emza) | vo |
| ئا | a | ئو | o | ئۇ(emza) | vu |
| ب | b | پ | p | ۋ | w |
| چ | c | ق | q | ش | x |
| د | d | ر | r | ي | y |
| ئې | e | س | s | ز | z |

THUYG-20 is the only Uyghur speech corpus freely available for download. It has been released by Xinjiang University and Tsinghua University jointly from China in 2015 [6], along with pre-built language models based on both word and morpheme units separately. THUYG-20 contains 23 hours of speech sampled at 16kHz from 371 speakers, transcripts of 12MB words in volume, and a vocabulary in the size of 45K [6]. Uyghur is a kind of agglutinative languages. A large number of words can be derived by concatenating different suffixes to one stem (or root), which leads to an explosion of the vocabulary, the data sparseness and out-of-vocabulary (OOV) problems while building a statistical language model [23]. For this issue, THUYG-20 offers recipes to build morpheme-based language models [6] and to perform morpheme-based decoding.

The phoneme set of THUYG-20, which is annotated with Latin letters, is shown in Table 1. It contains 32 phonemes in the standard phoneme set of Uyghur, including 8 vowels and 24 consonants, which are corresponding to the 32 letters in the Uyghur writing system[25]. However, it has to prefix an emza to a vowel when the vowel locates onset of a syllable, or appears independently, or successively follows another vowel in a word. The symbol emza just helps the input method of Uyghur to display a vowel letter in the correct form. It won't change the articulation of the vowel, and isn't contained in the standard Uyghur alphabet. As a result, an emza is prefixed to each vowel phoneme to produce a new marker for that vowel, in the phoneme set of THUYG-20. We use a "v" to represent an emza in Latin. As shown in Table 1, vA, vO, vU, va, ve, vi, vo, and vu are the eight symbols to represent the eight vowel phonemes each with an emza prefix. The eight vowel phonemes are also represented by A, O, U, a, e, i, o and u, correspondingly.

Based on the open-source platform Kaldi [20], it offers a complete recipe package to build up Uyghur ASR systems for THUYG-20 [6]. In Kaldi, a decoding graph is built in the form of a Weighted Finite State Transducer (WFST) [26][27]. The marker absent in Table 1, <eps>, which is inherited from the open-sourced library OpenFst[28], is set to represent an empty transition in a WFST. There are various kinds of silence frames in an utterance, for examples, at the beginning of a word, a sentence or a closure portion of a stop, which are all represented by "sil", as shown in Table 1.

## 3. An improvable GMM-HMM baseline system

Three GMM-HMM systems are trained successively: a monophone system, a triphone system, and a LDA+MLLT system, which have been also trained in the original work of THUYG-20 [6]. We reuse most of the settings, except the ones that are optimized. The systems all extract features of 13-dim Mel Frequency Cepstral Coefficients (MFCC), without energy information. The features are mean-value normalized by performing Cepstral Mean and Variance Normalization (CMVN) to overcome channel differences. The inputs of the monophone and triphone systems are all a single frame at each time step. However, in the LDA+MLLT system, a context window of 7 frames (3 on each side of the current frame) of features is used to exploit the neighboring information. The features from the context window are concatenated, and then performed dimension reduction to 40 by using Linear Discriminant analysis (LDA), and then Maximum Likelihood Linear Transform (MLLT). The alignments of training data of the monophone and triphone system are used to train the triphone and LDA+MLLT system, respectively. We build the LDA+MLLT system as the foundation to train a DNN-HMM hybrid system.

The GMM-HMM systems haven't been completely optimized in the recipes released along with THUYG-20 [6]. We find that the two parameters, --max-leaves and --tot-gauss, in the process of building a triphone decision tree, can be increased to make the limited training data more fully utilized. The optimization will offer a set of tied triphone HMM states with a more reasonable structure, and will also offer alignments of frame-level training labels in higher precision, for training neural network acoustic models.

## 4. Improving DNN-HMM systems using morpheme-based LMs

In a DNN-HMM hybrid system, the DNN is well trained to estimate posterior probabilities for the tied triphone HMM states [2], and the system performance is greatly improved by replacing GMMs with DNNs. However, THUYG-20 is too small to train very deep neural networks. To solve this problem, firstly, the GMM-HMM systems are ultimately optimized to prepare for training deeper and wider DNNs. Secondly, a morpheme-based decoder is independently developed for the DNN-HMM hybrid systems, which decreases WERs significantly.

### 4.1. Morpheme-based decoding with deeper and wider DNNs

The DNN acoustic model, in this paper, is a classical Feedforward neural network that has more than one hidden and fully connected layers. The hidden units use the sigmoid functions. The units of the softmax output layer are labeled with the tied triphone HMM states correspondingly. The input of the feedforward neural network is the concatenated fBank features from an 11-frame context window. The dimension of each fBank is 40, and the input layer has 440 units. The means of concatenated features are normalized by performing CMVN. After that, a training schedule is performed frame by frame to minimize the cross-entropy (xEnt) [2][29] between the labels and the predictions by the DNN acoustic model. The most significant optimization is that we add more hidden units to widen and add more hidden layers to deepen the feedforward neural network. Furtherly, the lattices and the frame-wise alignments generated by xEnt trained models are used as the starting point to perform the sequence-discriminative training using the criterion of Minimum Phone Error (MPE) [29]. As a result, the WERs decline.

Additionally, the WER will decrease significantly when the morpheme-based LMs are used in decoding, no matter which criterion, xEnt or MPE, is obeyed. When it is decoding with a morpheme-based LM, the morpheme sequence $\hat{m}$ is represented as

$$\hat{m} = \arg\max_{m} p(m \mid o) = \arg\max_{m} p(o \mid m) p(m) / p(o), \tag{1}$$

where $o = o_1 o_2 ... o_T$ is the input sequence of features, $m = m_1 m_2 ... m_n$ is the decoded morpheme sequence. $p(m)$ is the morpheme-based language model. $p(o \mid m)$ is a mixture of multiple components of various knowledge sources, where the acoustic model is the main component, and the triphone decision tree and the lexicon are both included. In addition, the alignments between different phonetic units in different scales are also included, such as the alignments between the HMM state sequence and the context-dependent phone sequence. Introducing the state sequence $q = q_1 q_2 ... q_T$ in a HMM model, where $q_t$ is the state at the time $t$, we can expand $p(o \mid m)$ as

$$p(o \mid m) = \sum_{q, m(q)} p(o, q, m(q) \mid m) \tag{2}$$

$$= \sum_{q, m(q)} p(o \mid q, m) p(q, m(q) \mid m) \tag{3}$$

$$= \sum_{\substack{q,cd(q),\\ ci(cd),m(ci)}} p(o \mid q,m)p(q,cd(q) \mid m)p(ci(cd) \mid cd,m)p(m(ci) \mid ci,m) \,, \tag{4}$$

where $m(q)$ represents the alignment between the morpheme sequence and the HMM state sequence, $cd(q)$ the alignment between the context-dependent phone sequence and the HMM state sequence, $ci(cd)$ the alignment between the context-independent phone sequence and the context-dependent phone sequence, and $m(ci)$ the alignment between the morpheme sequence and the context-independent phone sequence. $m(ci)$ also represents the morpheme-based lexicon corresponding to the transcript for LM building. We get

$$p(m(q)) = p(cd(q) \cap ci(cd) \cap m(ci)) \,. \tag{5}$$

The neural network acoustic model $p(o \mid q)$ is the main component of the mixture model $p(o \mid m)$, and

$$p(o \mid q) = \pi(q_0)\prod_{t=1}^{T} p(q_t \mid q_{t-1})\prod_{t=1}^{T} p(o_t \mid q_t) \,, \tag{6}$$

Where $\pi(q_0)$ and $p(q_t \mid q_{t-1})$ represent the initial state probabilities and the transition probabilities in the HMM respectively. $p(o_t \mid q_t)$ is the model to estimate the observation probabilities. As the DNN outputs the estimated posterior probabilities $p(q_t \mid o_t)$, the observation probability can be calculated as

$$p(o_t \mid q_t) = p(q_t \mid o_t)p(o_t) / p(q_t) \,. \tag{7}$$

In equation (7), $p(q_t)$ is the prior probability of each HMM state and can be estimated from the training set, and $p(o_t)$ is independent of the morpheme sequence and can be ignored.

We decode the optimal morpheme sequence according to equation (1), and convert it into the optimal word sequence. As shown by the experimental results in section 5, we have achieved state-of-the-art WER for the clean speech test task of THUYG-20, by using the DNN-HMM acoustic model trained with MPE criterion, and decoding with the morpheme-based LMs.

### 4.2 Decoding with morpheme-based LMs

#### 4.2.1 WFST-based decoding graphs

The WFST provides a promising approach to construct the static decoding graph in the scheme of state-of-the-art LVCSR, which simplifies the complex merging procedure of various knowledge sources into a series of basic operations on WFSTs[26][27][30]. The decoding scheme based on Viterbi beam search is also simplified. In this paper, we follow the definitions and the symbolic system in [27]. A weighted transducer is an 8-tuple $T = (\Sigma, \Delta, Q, i, F, E, \lambda, \rho)$ defined over a semiring $(\mathrm{K}, \oplus, \otimes, \bar{0}, \bar{1})$, where the symbols are specified the same as in [27], except that $i$ is used instead of $I$ to indicate the single initial state. The notations, including $p[e]$, $n[e]$, $i[e]$, $o[e]$ and $w[e]$, are also specified the same as in[27], except that $E_T[q]$ is specified as the set of transitions leaving state $q$ in $T$. The calculations involving $w[e]$, $\lambda$ and $\rho(f)$ are all defined over the semiring $(\mathrm{K}, \oplus, \otimes, \bar{0}, \bar{1})$. It is a standard way to configure the semiring as a tropical semiring [27], in the decoding scheme based on Viterbi beam search.

In a LVCSR system, four principal knowledge sources are represented by WFSTs, which are 1) the HMM transducer $H$, of which the input symbols are transition-ids and the output symbols are context-dependent phones; 2) the context-dependent phone transducer $C$, of which the input symbols are context-dependent phones, which generally are triphones, and the output symbols are context-independent phones; 3) the lexicon $L$, of which the input symbols are context-independent phones and the output symbols are words (or morphemes); 4) the N-Gram language model $G$, which is an acceptor taking words (or morphemes) both as the input and output symbols. The transition-ids are Kaldi-specific symbols[20], which uniformly encode the tied triphone HMM states and the transitions in the HMM topology. The tied triphone HMM states are also the targets to be classified by the DNN, which are encoded by pdf-ids [20]. The posterior probability of a tied triphone HMM state can be queried from the DNN's predictions based on the pdf-id. Each transition, in the HMM topology, is assigned a transition probability.

The four knowledge sources, which are represented by $H$, $C$, $L$ and $G$ respectively, are merged into a decoding graph [26], as shown in Kaldi's official document [31],

$$HCLG = asl(\min(rds(del(\overline{H} \circ \min(\det(C \circ \min(\det(L \circ G)))))))), \tag{8}$$

where *det* is specified as determinization, *min* as minimization, $\circ$ as composition, *rds* as remove-disambiguation-symbols, *asl* as add-self-loop, and $\overline{H}$ as an $H$ without self-loops. Generally, the inputs of the decoding graph $HCLG$ are transition-ids, and the outputs are words or morphemes. In addition, disambiguation symbols are introduced, such as #0, #1, #2, and #3, etc., to ensure that $HCLG$ can be determinized and minimized to be the least redundant.

#### 4.2.2 A morpheme lattice generating decoder for Uyghur DNN-HMM systems

Uyghur is a highly agglutinative language, with a large amount of derived words, which makes the vocabulary very large and leads to low lexical coverage [23]. To avoid this problem, morphemes are usually used as the units to build LMs. In the decoding procedure of THUYG-20, we build a 4-Gram morpheme-based LM $G_4$ with a size of 638MB. The static and fully composed decoding graph $HCLG_4$ is 16GB in size. However, more than 100GB memory is needed to perform the fully static determinization and minimization operations in the construction of $HCLG_4$. It isn't practical.

A basic strategy to use a LM as large as $G_4$ in decoding is to statically construct a fully composed decoding graph $HCLG_3$ by using a small LM $G_3$, and then to dynamically compose $HCLG_3$ with the difference of the large LM $G_4$ and the small LM $G_3$ [30][32]. $G_3$ is used to smear the LM scores along the weights of the graph $HCLG_3$, and the smeared weights will be updated by on-the-fly composition with $G_4$. $G_4$ is built by using the training text from THUYG-20. The transition probabilities, that are lower than 1e-5, are removed from $G_4$ to make the 3-Gram morpheme-based LM $G_3$. The difference of $G_4$ and $G_3$ is performed as $G_3^- \circ G_4$, where $G_3^-$ is the WFST by negating the weights of $G_3$. The decoder dynamically performs the ternary composition $HCLG_3 \circ G_3^- \circ G_4$, and generates on-the-fly partial of the decoding graph, and performs Viterbi searching and pruning to output a morpheme-based lattice. The ternary on-the-fly composition $HCLG_3 \circ G_3^- \circ G_4$ plays an equal role in decoding as the static and fully composed graph $HCLG_4$. However, the total size of $HCLG_3$, $G_3^-$ and $G_4$ is much smaller than the size of $HCLG_4$.

The BigLM decoding engine, which performs a Viterbi beam search with on-the-fly composition, has been used to construct a decoder, in Kaldi, named as gmm-latgen-biglm-faster for GMM-HMM systems. However, the BigLM decoder for DNN-HMM systems is missing. On the other hand, a BigLM decoder with the name of latgen-biglm-faster-mapped is used in the recipes of THUYG-20 [6]. State-of-the-art WER, by using the BigLM decoder and morpheme-based LMs for the DNN-HMM system, have been reported on the original publication of THUYG-20[6]. However, the source code and the application of the decoder were not released and have been missing since the time of the publication. So, the open resources from THUYG-20 don't support on-the-fly decoding for the DNN-HMM system with a very large morpheme-based LM.

Based on the platform of Kaldi, We have developed a decoder for the Uyghur DNN-HMM system independently, which performs the ternary on-the-fly composition $HCLG_3 \circ G_3^- \circ G_4$ in the process of decoding Uyghur speeches. So, we call the decoder a Morpheme Lattice Dynamically Generating Decoder for Uyghur DNN-HMM systems (MLDG-Decoder)[4].

### 4.2.3 On-the-fly composition with FEBABOS

When the MLDG-Decoder decodes Uyghur speeches by using the very large morpheme-based LM $G_4$, the strategy of the ternary on-the-fly composition $HCLG_3 \circ G_3^- \circ G_4$ not only achieves the performance comparable to that of using the static and fully composed decoding graph $HCLG_4$, but also maintains the technical characteristics of one-pass decoding and low memory consuming. In the case of back-off n-Gram language models, the algorithm of the ternary on-the-fly composition are significantly different from the standard composition algorithm of WFSTs[27].

In the standard composition algorithm of WFSTs, the back-off transitions with input symbols of epsilons normally take part in matching of input and output symbols, and composing of weights, which competes with non-back-off transitions. After that, the shortest path wins from numerous candidates by using a filter transducer[29]. It has been guaranteed that all back-off paths always score worse than their corresponding non-back-off paths. However, the worst path will always win if the ternary on-the-fly composition $HCLG_3 \circ G_3^- \circ G_4$ was performed by fully complying with the standard composition algorithm. In $G_3^-$, the weights are negated according to $G_3$. As a result, the worst path in $G_3$, which is always a back-off epsilon path, results to be the best one in $G_3^-$. The negated weights make the decoder always output the worst back-off path. To avoid this misjudgment, the MLDG-Decoder adopts a strategy called "Forbidden Epsilons But Activated by Back-Off States (FEBABOS)". We named the ternary on-the-fly composition algorithm that complying with this strategy as "on-the-fly composition with FEBABOS", and the pseudocode of the algorithm is shown in Fig 1.

The algorithm takes $HCLG_3$, $G_3^-$, $G_4$ and $S_{last}$ as input, and outputs $S_{curr}$. $S_{last}$ and $S_{curr}$ are both a list of tokens for the last and the current frame of speech signals respectively. The results of on-the-fly composition are recorded in the list of tokens, where each token takes a 3-tuple $(q_1,(q_2,q_3))$ of states as a new state and takes the total cost along the path that arrived the new

[4] https://github.com/studyself/kaldi

**Algorithm 1** On-the-fly Composition with Forbidden Epsilons But Activated By Back-Off States

**Input:** $HCLG_3, G_3^-, G_4, S_{last}$
**Ouput:** $S_{curr}$

```
S_curr ← ∅
while S_last ≠ ∅ do
    [(q_1, (q_2, q_3)), cost] ← HEAD(S_last)
    DEQUEUE(S_last)
    for each e_1 ∈ E_{HCLG_3}[q_1] such that i[e_1] ≠ ϵ do
        if o[e_1] = ϵ then
            q'_1 ← n[e_1]
            cost ← cost ⊗ w[e_1] ⊗ Acoustic(i[e_1])
            ENQUEUE(S_curr, [(q'_1, (q_2, q_3)), cost])
            continue
        else
            GW_4 ← 1̄
            GetArc3:
            if find e_2 ∈ E_{G_3^-}[q_2] such that i[e_2] = o[e_1] then
                if o[e_2] = ϵ then
                    q'_1 ← n[e_1], q'_2 ← n[e_2]
                    GW_4 ← GW_4 ⊗ w[e_1] ⊗ w[e_2]
                    cost ← cost ⊗ GW_4 ⊗ Acoustic(i[e_1])
                    ENQUEUE(S_curr, [(q'_1, (q'_2, q_3)), cost])
                    continue
                else
                    GetArc4:
                    if find e_3 ∈ E_{G_4}[q_3] such that i[e_3] = o[e_2] then
                        q'_1 ← n[e_1], q'_2 ← n[e_2], q'_3 ← n[e_3]
                        GW_4 ← GW_4 ⊗ w[e_1] ⊗ w[e_2] ⊗ w[e_3]
                        cost ← cost ⊗ GW_4 ⊗ Acoustic(i[e_1])
                        ENQUEUE(S_curr, [(q'_1, (q'_2, q'_3)), cost])
                        continue
                    else
                        Get e_backoff4 ∈ E_{G_4}[q_3] such that i[e_backoff4] = ϵ
                        q_3 ← n[e_backoff4], GW_4 ← GW_4 ⊗ w[e_backoff4]
                        go to GetArc4
                    end if
                end if
            else
                Get e_backoff3 ∈ E_{G_3^-}[q_2] such that i[e_backoff3] = ϵ
                q_2 ← n[e_backoff3], GW_4 ← GW_4 ⊗ w[e_backoff3]
                go to GetArc3
            end if
        end if
    end for
end while
return S_curr
```

Fig. 1 The pseudocode of on-the-fly composition with FEBABOS

state as a new cost. Suppose that $Q_1$, $Q_2$, $Q_3$ and $E_1$ $E_2$ $E_3$ are the finite sets of states and transitions of $HCLG_3$, $G_3^-$ and $G_4$ respectively, where we have $q_1 \in Q_1$, $q_2 \in Q_2$, $q_3 \in Q_3$ and $e_1 \in E_1$, $e_2 \in E_2$, $e_3 \in E_3$.

Starting from the currently arrived states in $HCLG_3$, the MLDG-Decoder searches around all the candidate paths, along which the transitions, corresponding to the current frame, propagate forward in the ternary on-the-fly composed graph $HCLG_3 \circ G_3^- \circ G_4$, while the total cost along the path is being updated. The algorithm only deals with the transitions of non-epsilon inputs, which indicates emitting probabilities along with the transitions (line 5). The MLDG-Decoder won't search for transitions in $G_3^-$ and $G_4$ to matching the transitions with epsilon outputs in $HCLG_3$ and $G_3^-$ respectively, but ends the dynamic composition immediately (lines 6-10 and lines 15-20). In one word, it is forbidden to match the transitions with epsilon output labels. However, $G_3^-$ and $G_4$ are not treated as epsilon-free, as the back-off states and transitions with epsilon input labels will act as a relay station. When $o[e_1]$ (or $o[e_2]$) is non-epsilon, and it's failed to find a transition to match it as $i[e_2] = o[e_1]$ (or $i[e_3] = o[e_2]$), the MLDG-decoder will pass by the back-off state through a transition of an epsilon input label, and continue to complete the matching (lines 35-38 and lines 29-32). If the MLDG-decoder find transitions to match the non-epsilon output labels $o[e_1]$ and $o[e_2]$, as $i[e_2] = o[e_1]$ and $i[e_3] = o[e_2]$, on-the-fly composition will be performed the same as the standard static composition (lines 23-28). In general, the back-off states and epsilon transitions in $G_3^-$ and $G_4$ are forbidden to be matched and composed in the process of on-the-fly composition, but will be activated and passed by when the non-epsilon matching fails.

$Acoustic(i[e_1])$ represents the negative log-likelihood for the current frame given a non-epsilon input label (lines 8, 18 and 26), which is a transition-id. The negative log-likelihood is performed to satisfy the condition of the tropical semiring. $GW_4$ represents the Graph Weights to be updated. Additionally, the new tokens generated with the same 3-tuple $(q_1,(q_2,q_3))$, are all pushed into the list $S_{curr}$ before they are combined by using the $\oplus$-operation. We take combining and pruning of the paths, and generating of the morpheme-based lattice as the next step after on-the-fly composition.

ئىشچى ئىشتىن چۈشتى

Words: vixci vixtin cUxti

Morphemes: vix ci vix tin cUx ti

English: The Worker is off work

ئىشتىن چۈشكەن ئىشچى

Words: vixtin cUxkAn vixci

Morphemes: vix tin cUx kAn vix ci

English: The off-work worker

(a)

vix ci vix tin cUx ti

vix tin cUx kAn vix ci

(b)

Fig 2. A mini morpheme-based corpus in Uyghur, where in (a) words are split into morphemes, and (b) is the mini corpus.

A 2-Gram morpheme-based LM is constructed in Uyghur based on a mini corpus as shown in Fig. 2, to explain the searching strategy performed in $G_3^-$ or $G_4$ by the algorithm of on-the-fly composition with FEBABOS. It is a statistical LM represented by a WFST as shown in Fig 3. The LM is constructed by following the standard process specified in the Kaldi recipes. First, a 2-Gram language model is built in ARPA format based on the mini morpheme-based corpus by using SRILM. Second, the ARPA LM is converted into a WFST and all epsilon input labels of back-off transitions are converted to the disambiguation symbol #0. Third, the special symbols of beginning and end of sentences, <s> and <\s>, are replaced by epsilons. At last, the graph is simplified by performing epsilon removing operations. As shown in Fig. 3, the algorithm 1 will walk along the path which is in bold lines: $0 \to 1 \to 7 \to 3 \to 1 \to 2$ , as it will performs in $G_3^-$ or $G_4$, if the current state is the state 0 and the input morpheme sequence is “tin cUx vix”. The path in bold lines shows that the back-off state 1, the back-off transition $0 \to 1$ and $3 \to 1$ are always passed by during on-the-fly composition.

The idea of on-the-fly composition has been firstly proposed by Dolfing etc. [32] to deal with very large n-Gram LMs. Hori etc. [30] have re-implemented the algorithm and developed it into a so-called on-the-fly rescoring approach. Kaldi also has implemented a decoding engine named as BigLM in its open-sourced code [33]. These implementations are all in the way of on-the-fly composition, however they still have many differences. First, the implementation in Kaldi is open-sourced, while it is only outlined in principle in [30] and [32]. Second, in the versions of [30] and [32], the composition of $F = G_3^- \circ G_4$ is performed in advance, and then on-the-fly composition of $HCLG_3 \circ F$ follows. However, the ternary compositions actually are dynamically performed at the same time, in Kaldi. As shown in Fig. 1, in the compositions of $HCLG_3 \circ G_3^-$ and $G_3^- \circ G_4$, the forbidding and activating of back-off states and transitions are all performed in the same while-loop. We don't think it's completely correct to take $G_3^-$ and $G_4$ as epsilon-free. Third, in [30] and [32], the way to deal with back-off states and transitions is not explained as clear as the code from Kaldi. They only describe the algorithm on a top-level, and give examples partially cover the details in Algorithm 1. MLDG-Decoder is an extension of the BigLM engine, which can be used for Uyghur DNN-HMM systems.

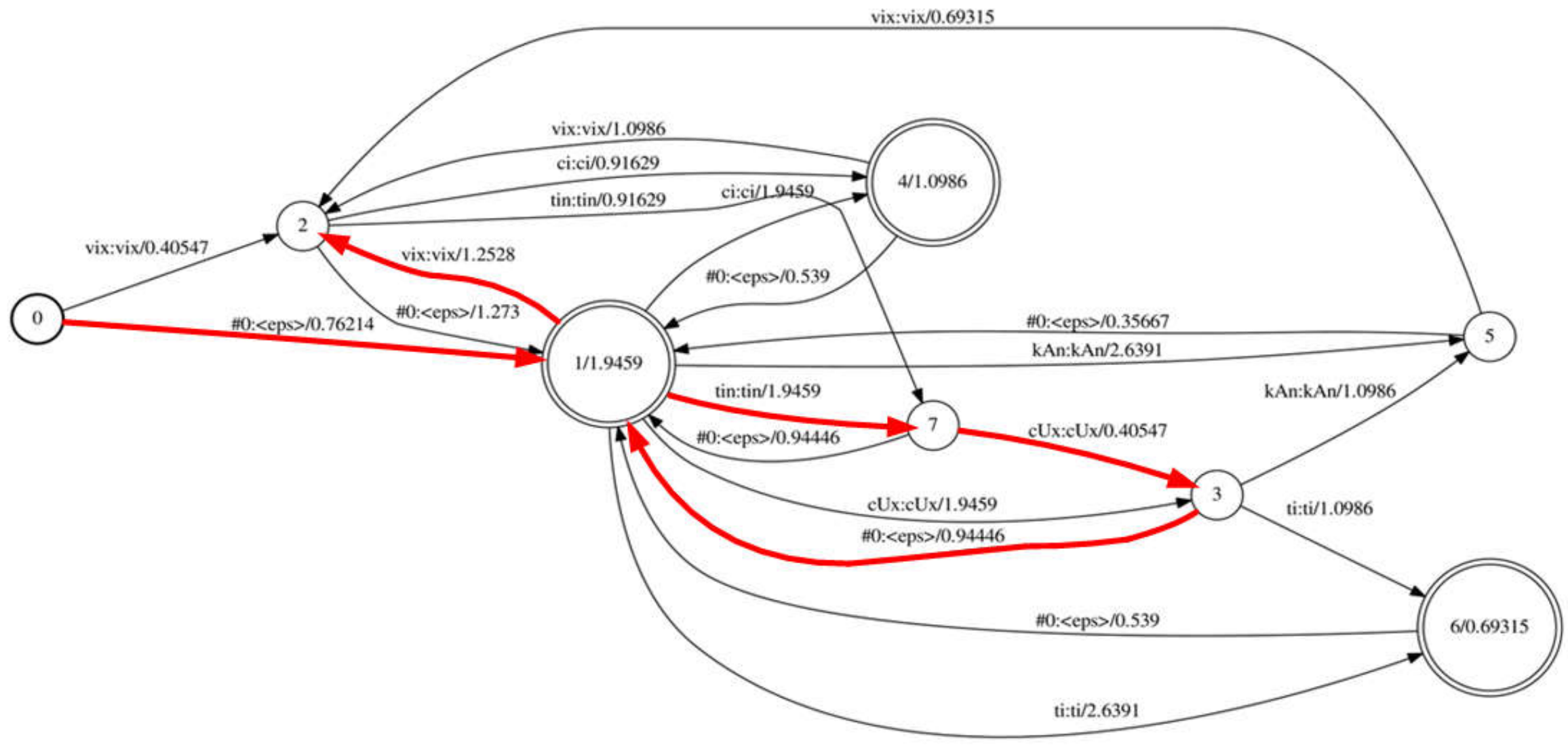

Fig 3. A 2-Gram morpheme-based LM of Uyghur in the form of a WFST

## 5. Experimental results and discussions

### 5.1. Optimization of the GMM-HMM baseline systems

We have conducted a series of experiments to evaluate the GMM-HMM systems, which have been introduced in Section **3**. The two parameters, --max-leaves and --tot-gauss, are investigated during building a triphone decision tree. We increase the maximum number of leaves to make the decision tree split more completely as long as the training data allows. We also increase the total number of Gaussian functions to improve the fitting accuracy of the emission probabilities $p(o_t \mid q_t)$, as shown in Equation(7). After that, iterations should be increased to ensure the GMM-HMM training process can be converged. In our last paper[12], similar experiments have been performed in advance, however, the training epochs didn't increase to the point that the models can fully converge.

A triphone system and a LDA+MLLT system are trained in the experiments, and set –max-leaves in the range of 2,500~5,000, and –tot-gauss in the range of 10,000~50,000, respectively. Then, the Uyghur speeches for test are decoded by using the word-based LM and the morpheme-based

LMs respectively. As shown in Fig. 4, the WERs decline significantly as –tot-gauss increases. The triphone and LDA+MLLT systems are also improved when –max-leaves increase from 2,000 to 4,000 and from 2,500 to 4,500, respectively. The morpheme-based decoding outperforms the word-based decoding both for the two systems. To be noted that, the training data will be inadequate when –max-leaves is larger than 4,500, and the improvement will be little when –tot-gauss is larger than 35,000. The results before and after the optimization are shown in Table 2 and Table 3, respectively for the triphone and LDA+MLLT systems.

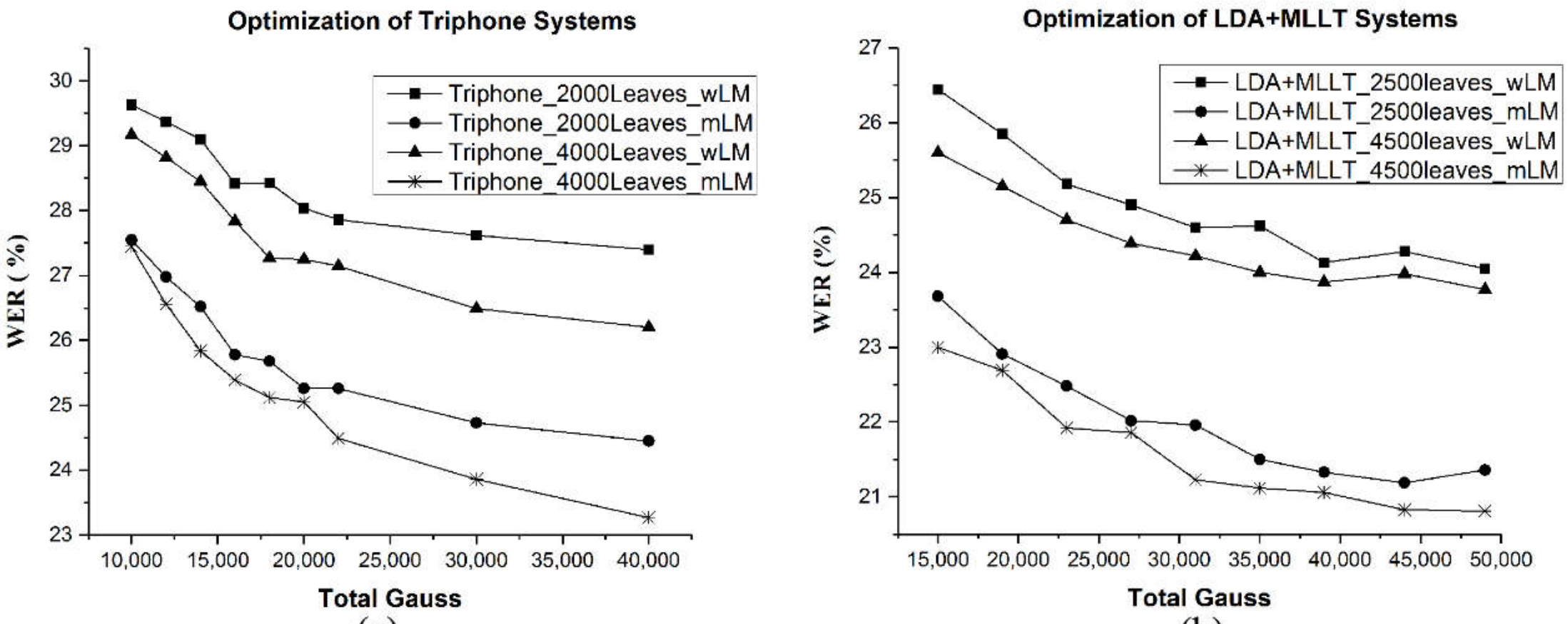

(a) (b)

Fig 4. It decreases WERs by increasing the maximum number of leaves and the total number of Gaussian functions, where (a) shows the results of the triphone system, and (b) shows the results of the LDA+MLLT system. 4500leaves means –max-leaves in the value of 4500, and wLM means decoding with a 3-Gram word-based LM, and mLM means BigLM decoding with morpheme-based LMs.

**Table 2 Results of triphone systems**

| LM | WER（%） | |
|---|---|---|
| | Before optimization max-leaves=2000, tot-gauss=10000 | After optimization max-leaves=4000, tot-gauss=30000 |
| Word-based | 29.98 | 26.29 |
| Morpheme-based | 27.51 | 23.97 |

According to the results above, we set –max-leaves 4,000 and –tot-gauss 30,000 to train a triphone system. Based on the alignments from the triphone system, we set –max-leaves 4,500 and –tot-gauss 35,000 to train a LDA+MLLT system. During the training process, a data driven phonetic decision tree is re-built, and 3504 tied triphone HMM states are produced, which correspond to the leaves of the tree. All non-silent triphones use a 3-state HMM topology, while silent triphones use a 5-state HMM topology. The results before and after the optimization are shown in Table 2 and Table 3, respectively for the triphone and LDA+MLLT systems. In Table 2 and Table 3, the parameters specified as ”Before optimization” are also the configuration used in the original recipes released along with [6]. The set of tied triphone HMM states and the frame-level alignments are both inherited from the LDA+MLLT system, and will be reused in the successive training of DNN-HMM systems.

**Table 3 Results of LDA+MLLT systems**

| LM | WER（%） | |
|---|---|---|
| | Before optimization max-leaves=2500, tot-gauss=15000 | After optimization max-leaves=4500, tot-gauss=35000 |
| Word-based | 26.47 | 24.08 |
| Morpheme-based | 23.91 | 21.02 |

### 5.2. Optimization of DNN-HMM systems and morpheme-based decoding

Thanks to the improved set of targets and the improved alignments by ultimate tuning of the GMM-HMM systems, the DNN-HMM hybrid system can try deeper and wider neural networks in spite of the small volume of training data in THUYG-20. We have trained several DNN acoustic models with 5, 6, 7 and 8 hidden layers respectively. The hidden layers in DNNs are set to 1024 or 2048 in dimension. As it has got 3,504 tied triphone HMM states correspond to the leaves of the phonetic decision tree, the output layer of all the DNNs are set to 3,504 in dimension.

The deep feedforward neural network is trained, frame by frame, by using SGD (Stochastic Gradient Descent) to minimize the cross-entropy between the labels and the predictions by the network. The mini-batch training of 256 frames is performed. The learning rate is set to 0.008 initially, and is halved when the frame accuracy improved less than 0.5% between two continuing epochs on the cross-validation set. When the improvement is less than 0.1%, the training process terminates.

Based on the alignments and morpheme lattices generated by the xEnt training, the sequence-discriminative training with MPE as the objective function is performed. The learning rate is set to 1e-5 constantly and lattices are regenerated after each epoch. The training process won’t stop until the WER converges. In the recipes and test results, that have been released along with THUYG-20 [6], the MPE training converged only after 3 epochs.

However, in our experiments, the MPE training can last up to 40 epochs at most, in the same training scheme of the same learning rate. It indicates that the sequence-discriminative training with the MPE criterion has got more room for improvement from the series of optimization on GMM-HMM and DNN-HMM systems, and also from the usage of the MLDG-Decoder.

The experimental results show that the DNN acoustic model with 7 hidden layers and 2048 nodes per hidden layer outperforms the others. The DNN acoustic models trained with MPE criterion outperform the ones trained with xEnt criterion. When the acoustic models are the same, the morpheme-based decoding results are better than the word-based decoding results.

The WERs by decoding with the word-based LM is shown in Table 4, where THUYG-20 indicates the best results from the original publication [6]. It has trained a DNN acoustic model with only 4 hidden layers in [6], just due to the insufficient optimization of the training process and the inadequate training data in THUYG-20. As shown in Table 4, the WERs don't decline much after optimization, whether the system is trained based on the xEnt or the MPE criterion. It is believed that the word-based LM constitutes the bottleneck of the decoding scheme, although the deeper and wider network has improved the DNN-HMM acoustic model.

**Table 4 Results of DNN-HMM systems with a word-based LM**

| WER（%） | THUYG-20<br>4 hidden layers，1024 per layer | After optimization<br>7 hidden layers，2048 per layer |
|---|---|---|
| DNN (xEnt) | 19.57 | 19.55 |
| DNN (MPE) | 18.95 | 18.45 |

Three decoding strategies have been arranged for the morpheme-based LMs. First, MLDG-Decoder, which is based on on-the-fly composition with FEBABOS, is used, as shown in Section 4.2.3. Second, the two-pass decoding based on rescoring is performed. The decoding graph $HCLG_3$ is built statically in the first pass, and the pruning parameters are set to the same as that of MLDG-Decoder. The first pass generates a morpheme lattice containing multiple hypotheses. In the second pass, the morpheme lattice is rescored by using the much larger morpheme-based LM $G_4$ to output the best hypothesis. Third, the static and fully composed decoding graph $HCLG_4$ is constructed directly, and a static decoding on $HCLG_4$ is performed to output a morpheme lattice.

**Table 5 Results of DNN-HMM systems with morpheme-based LMs**

| WER（%） | THUYG-20,<br>4 hidden layers ，<br>1024 per layer | After optimization：7 hidden layers，2048 per layer | | |
|---|---|---|---|---|
| | | MLDG-Decoder | Rescoring | $HCLG_4$ |
| DNN (xEnt) | 17.40 | 15.96 | 17.06 | 15.46 |
| DNN (MPE) | 16.58 | **14.54** | 15.35 | **14.24** |

The WERs by decoding with the morpheme-based LMs is shown in Table 5. As shown in Table 5, the results after optimization are generally better than that from the original publication [6], which has been dominant on THUYG-20 as state-of-the-art results since the publication. When the DNN acoustic model is trained by using the MPE criterion, the result, by decoding on the static and fully composed graph $HCLG_4$, decreases 2.34% absolutely, compared with the BigLM decoding result of MPE from the THUYG-20 original publication [6]. The relative decrease is 15.11%. However, more than 100GB memory is needed to construct the static decoding graph $HCLG_4$. It isn't practical. We have upgraded the hardware several times to meet the experimental conditions. On the other hand, the MLDG-Decoder achieves a comparable WER while the memory consuming is not high. Compared with the BigLM decoding result from the THUYG-20 original publication [6], the WER decrease 2.04% absolutely and 12.3% relatively. In comparison, the rescoring strategy performs worst, of which the WERs are worse than that of MLDG-Decoder by about 1% absolutely. In terms of execution efficiency, the Real Time Factor (RTF) of the MLDG-Decoder is 1.48. The RTF of the first pass of the rescoring strategy is 1.53. The RTF of the fully static decoding based on $HCLG_4$ is 0.79. Generally, the decoding scheme of the MLDG-Decoder is practical and achieves significant decrease of WERs on THUYG-20.

## 6. Conclusions

The ultimate optimization of the phonetic tree building process and GMM-HMM systems training, helps to generate a set of tied triphone HMM states with a more reasonable structure and alignments of frame-level training labels in higher precision, respectively. The above optimization enables us to successfully train a DNN-HMM acoustic model with 7 hidden layers and 2048 nodes in each layer, while the DNN has only 4 layers and 1024 nodes in each layer in the original publication of THUYG-20. Both of xEnt and MPE criterions are used to train DNN-HMM acoustic models. The experimental results show that sequence-discriminative training outperforms the frame-wise xEnt training.

Another bottleneck of the Uyghur ASR system performance, is the word-based decoder. The morpheme-based decoder for the DNN-HMM system has been missing since the first release and publication of THUYG-20. The article firstly discloses a morpheme-based decoder, named as MLDG-Decoder, which decodes for the DNN-HMM hybrid systems with morpheme-based LMs. The decoder employs an algorithm, named as on-the-fly composition with FEBABOS, which successfully decreases the WERs and makes the memory consumption acceptable. Three morpheme-based decoding strategies are experimented for the DNN-HMM hybrid systems: using MLDG-Decoder, the two-pass decoding based on rescoring, and the fully static and morpheme-based decoding. The results of decoding on the static and fully composed graph refresh state-of-the-art WER on the clean and noiseless speech test task to 14.24%, which declines 2.34% absolutely. However, it consumes more than 100GB memory to build the static graph. The MLDG-Decoder achieves a comparable WER of 14.54%, which decreases 12.3% relatively compared with the BigLM decoding result from the THUYG-20 original publication. The MLDG-Decoder's memory consumption is acceptable.

Before our work, state-of-the-art WER on the clean and noiseless speech test task of THUYG-20 hasn't been refreshed since the first release and publication, which shows that the development of Uyghur ASR has lagged far behind that of the mainstream languages. It is reasonable to take our work as a new starting point to accelerate Uyghur ASR's development.